\documentclass[10pt,twocolumn,letterpaper]{article}

\usepackage[pagenumbers]{cvpr} 

\definecolor{cvprblue}{rgb}{0.21,0.49,0.74}
\usepackage[pagebackref,breaklinks,colorlinks,allcolors=cvprblue]{hyperref}
\usepackage{graphicx}
\usepackage{multirow}
\usepackage{amssymb}
\usepackage{pifont}
\newcommand{\xmark}{\ding{55}}%

\def\eg{\emph{e.g.}} 
\def\ie{\emph{i.e.}}

\def\paperID{11258} 
\def\confName{CVPR}
\def\confYear{2025}

\title{HyperNVD: Accelerating Neural Video Decomposition via Hypernetworks}

\author{
Maria Pilligua$^{1,2,}$\footnotemark[1]\quad
Danna Xue$^{1,2,}$\footnotemark[1]~~\footnotemark[2]\quad 
Javier Vazquez-Corral$^{1,2}$\\
$^1$ Universitat Autònoma de Barcelona \quad $^2$ Computer Vision Center\\
{\tt\small \{mpilligua, dxue, javier.vazquez\}@cvc.uab.cat}
}

\begin{document}
\maketitle

\renewcommand{\thefootnote}{\fnsymbol{footnote}}
\footnotetext[1]{These authors contributed equally to this work.}
\footnotetext[2]{Corresponding author.}

\begin{abstract}
Decomposing a video into a layer-based representation is crucial for easy video editing for the creative industries, as it enables independent editing of specific layers. Existing video-layer decomposition models rely on implicit neural representations (INRs) trained independently for each video, making the process time-consuming when applied to new videos. Noticing this limitation, we propose a meta-learning strategy to learn a generic video decomposition model to speed up the training on new videos. Our model is based on a hypernetwork architecture which, given a video-encoder embedding, generates the parameters for a compact INR-based neural video decomposition model. Our strategy mitigates the problem of single-video overfitting and, importantly, shortens the convergence of video decomposition on new, unseen videos. Our code is available at: \href{https://hypernvd.github.io/}{\textcolor{Bittersweet}{https://hypernvd.github.io/}}.
\end{abstract}


\section{Introduction}
\label{sec:intro}

Video editing plays a crucial role in today's creative industries, from film-making and advertising to social media content creation. It enables creators to bring their visions to life, enhancing storytelling through visual effects, scene adjustments, and content personalization. However, most current video editing techniques face several challenges: they require extensive manual intervention~\cite{yu2023videodoodles}, lack consistent control over edits~\cite{cong2023flatten, geyer2023tokenflow}, and are time-consuming~\cite{nam2022neural, chan2023hashing}, limiting their scalability for real-world applications.

Current video editing workflows in professional software resemble traditional 2D animation. Creators import video into editing software~\cite{AdobeAE, DaVinci} and edit in overlaid layers for each video frame. While motion tracking tools can interpolate adjustments between keyframes~\cite{GeoTracker, nuke}, handling occlusions and ensuring consistency remain challenging since tracking errors make editing cumbersome. These problems highlight the importance of accurate layered representation and frame mappings.

\begin{figure}[t]
  \centering
   \includegraphics[width=\linewidth]{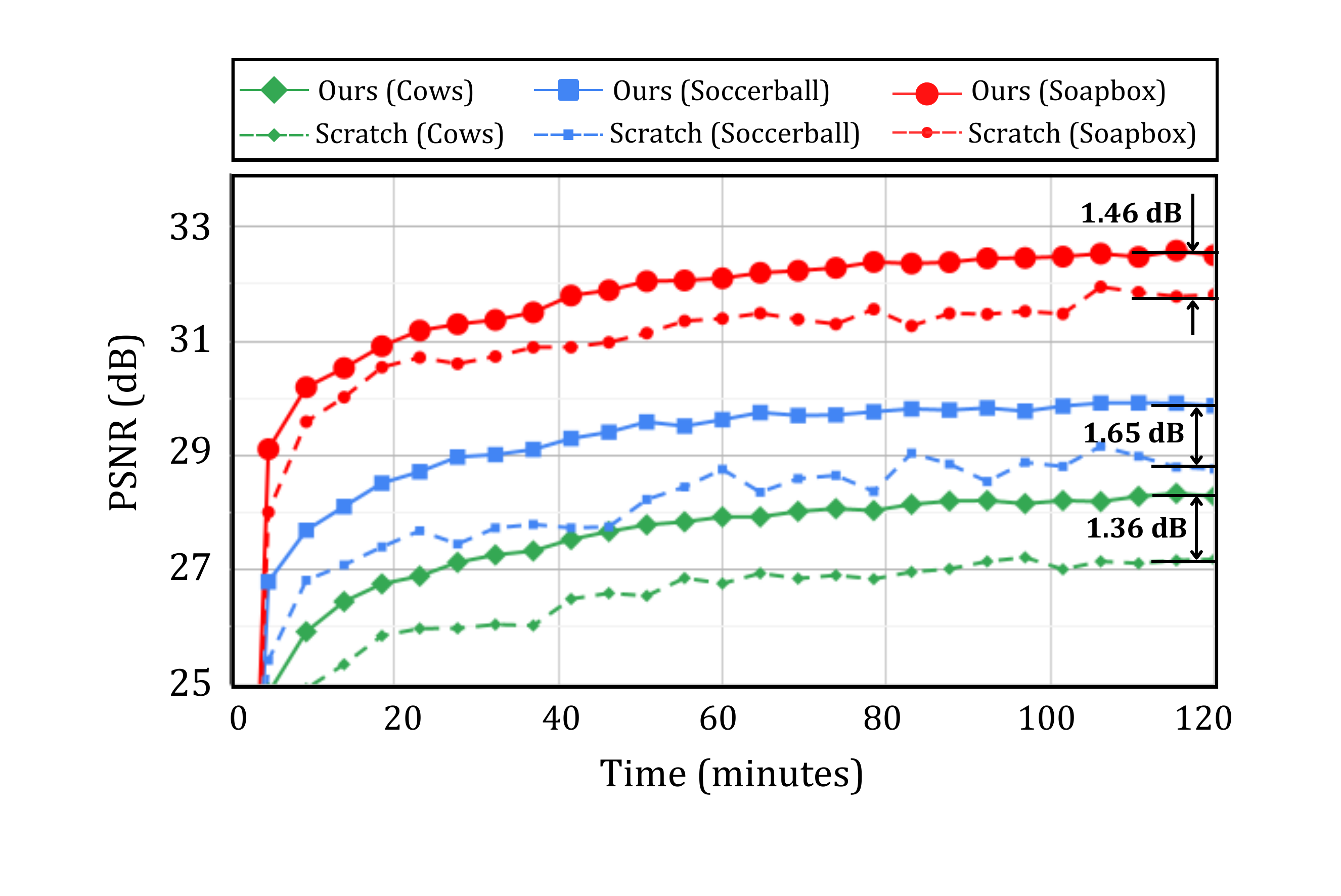}
   \vspace{-6mm}
   \caption{Fine-tuning the video decomposition model from our metamodel (HyperNVD) versus training from scratch on unseen videos shows clear advantages. With initialization from our HyperNVD trained on 15 videos, the model converges faster to the same PSNR and ultimately achieves better performance.}
   \label{fig:teaser}
    \vspace{-5mm}
\end{figure}

Layer-based video decomposition methods~\cite{smirnov2021marionette, ye2022deformable, nam2022neural, chan2023hashing, lee2024rna} represent video with several texture layers, where each layer consistently represents specific video content (\eg, moving objects or backgrounds). These approaches are mainly based on compact implicit neural representations (INRs), which encode each video individually. The model projects pixel coordinates and frame indices into a canonical 2D texture space, and then, it maps to RGB values to reconstruct each frame. In this way, it is possible to apply simple edits on the canonical texture layer and then propagate them to the entire video, greatly relieving the need for manual adjustments. Furthermore, the parameter sizes and training times are reduced compared to training a generalized video object segmentation model~\cite{athar2023tarvis} and global flow estimation model~\cite{doersch2022tap, karaev2023cotracker}. However, due to the nature of INR, the neural video decomposition model lacks generalization and requires tens of minutes to train a 480p video~\cite{chan2023hashing}. 

In this paper, we propose a meta-learning strategy in layer-based video decomposition to speed up the training. More in detail, we introduce a hypernetwork (hypernet) that, based on the video embedding predicted by VideoMAE~\cite{tong2022videomae}, generates the parameters for a compact INR model. Our strategy facilitates the generalization to more than one video and also allows for speed-up training on new, unseen videos, as we can use the parameters given by the hypernet to initialize the neural video decomposition model for further fine-tuning. Figure~\ref{fig:teaser} shows three examples of this speed-up. In this figure, we compare the training time and the PSNR obtained for training an unseen video starting from the parameters suggested by our hypernet or from a random parameter set. Our method obtains the same PSNR with more than 30 minutes less.

\vspace{2mm}
\noindent \textbf{Contributions. }Our main contributions are as follows:

\begin{itemize}
    \item We incorporate a meta-learning strategy to learn a generic video decomposition method. Our strategy consists of learning a hypernet that, given a video-encoder embedding, predicts the parameters for an INR-based video decomposition model. 
    \item Our framework relieves the problem of overfitting on a single video and shortens the convergence time of the neural video decomposition model on unseen videos.
    \item The experimental results show the generalizability and appealing video editing ability of our approach.
\end{itemize}

\section{Related Work}
\label{sec:related}

\subsection{Layered video decomposition}
In image editing and animation, an image is often separated into multiple layers, each with color and opacity~\cite{sengupta2020background, xu2017deep}, allowing independent adjustments before stacking them for the final result. This layered representation is also applied to video editing for simplifying the workflow and ensuring consistency~\cite{wang1994representing, jojic2001learning, rav2008unwrap}. Neural video decomposition approaches~\cite{kasten2021layered, chan2023hashing, ye2022deformable} decompose videos into consistent 2D texture layers across frames and are usually trained in a self-supervised manner using motion and appearance constraints. LNA~\cite{kasten2021layered} uses implicit neural representations to reconstruct videos via neural atlases and alpha blending. While encoding each video individually reduces parameter size and rendering time, reconstructing high-frequency details and long training time remain challenges. Omnimatte~\cite{lu2021omnimatte} and Factormatte~\cite{gu2023factormatte} add appearance priors to model associated effects like shadows, or reflections among others. Hashing-nvd~\cite{chan2023hashing} and CoDeF~\cite{ouyang2024codef} introduce hash encoding to accelerate optimization. However, these models still need more than 40 minutes of training on one single video~\cite{chan2023hashing}. We propose to use a hypernet to generate parameters of the video decomposition INR. In this way, we can further shorten the training time of any new, unseen video by initializing its training from the weights given by the hypernet.

\subsection{Hypernetwork}
Hypernetworks (or hypernets)~\cite{ha2016hypernetworks, chauhan2024brief} is a technique that enhances the adaptability and flexibility of deep learning models. Instead of learning task-specific models with fixed parameters, a hypernet dynamically generates parameters for a target model based on task~\cite{Oswald2020Continual, finn2017model, Nichol2018OnFM}, or data~\cite{su2020blindly, kim2023generalizable, maiya2024latent, sen2024hyp}. By learning a shared, adaptable representation, hypernets help target models generalize better across different conditions. Once trained, they use embeddings as input to modulate the target model’s weights and reduce optimization times on unseen tasks~\cite{finn2017model, Nichol2018OnFM}. Recent work has further improved input-specific model flexibility, especially in INR applications~\cite{kim2023generalizable, chiang2022stylizing, maiya2024latent}. The work most closely related to ours is HyP-NeRF~\cite{sen2024hyp}, which employs a hypernet to generate instance-specific parameters for hash encoding and NeRF, linking text to 3D object reconstructions. This framework relates semantic similarities across models for different instances but focuses on simple 3D objects with limited shape and color variations. In contrast, our video modeling task is more complex and requires attention to multiple objects, detailed motion, and varied appearances.

\section{Accelerate Neural Video Decomposition via Hypernetwork}
\label{sec:method}

\subsection{Framework overview}
\begin{figure*}[ht]
  \centering
   \includegraphics[width=0.95\linewidth]{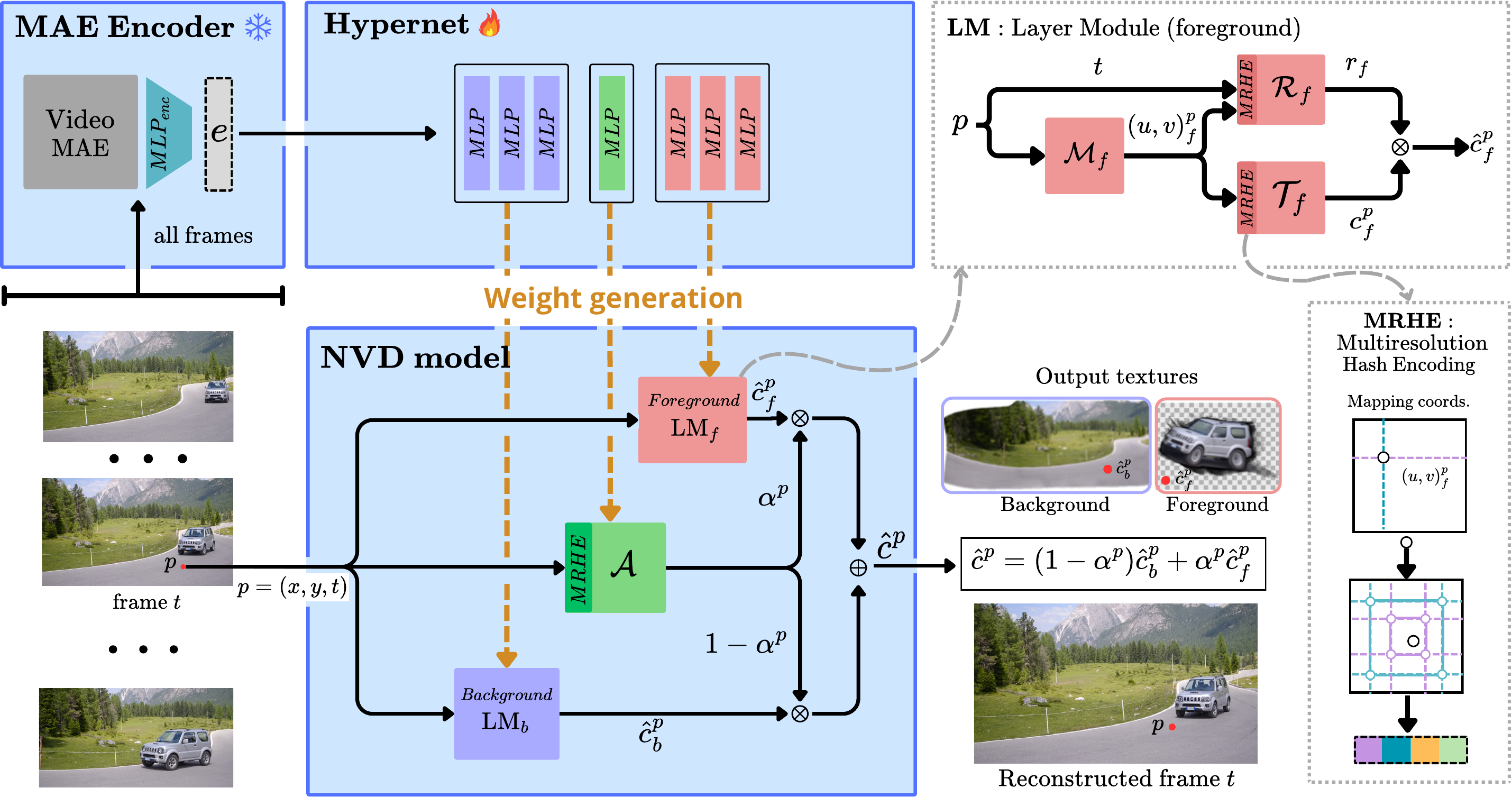}
  \caption{The architecture of our HyperNVD. HyperNVD consists of \textit{i) the MAE encoder to generate video embedding}, \textit{ii) the hypernet to generate model parameters}, and \textit{iii) the target neural video decomposition (NVD) model}. Given an input video, the MAE encoder encodes it to a compact embedding $e$. Then the hypernet $\mathcal{H}$ generates the parameters of the NVD model, including the MultiResolution Hash Encoding (MRHE) and the model weights. The NVD model includes two layer modules (foreground and background) for reconstructing different components in the video and an alpha module for predicting the soft mask to blend the layers. The reconstructed frame is generated by adding the foreground and background texture layers with an opacity map. Each layer module consists of a mapping module, a texture module, and a residual module.}
  \label{fig:framework}
  \vspace{-2mm}
\end{figure*}

Our goal is to learn a generalizable model for neural video decomposition that helps speed up the optimization of unseen videos. Our HyperNVD consists of three main components: i) \textit{a video embedding generation model}, ii) \textit{a hypernet to generate model parameters}, and iii) \textit{a neural video decomposition (NVD) model}. Figure~\ref{fig:framework} shows an overview of our model.
First, we compress the features predicted by a pre-trained video transformer, VideoMAE~\cite{tong2022videomae}, and use the compressed embedding as input to the hypernet. Then the hypernet predicts the weights and the multiresolution hash encoding parameters of the target NVD model. 
Our architecture allows the training of multiple videos simultaneously, which enables fast adaptation of unseen videos. This fast adaptation is accomplished by considering the parameters given by the hypernet for further fine-tuning.

\vspace{2mm}
\noindent \textbf{Neural video decomposition (NVD) model.} 
Standard layered video decomposition models based on INR~\cite{chan2023hashing} work by taking video coordinates and frame indices $p=(x,y,t)$ as input and decomposing the video into a background layer and a foreground layer. 

As shown in the Layer Module of Figure \ref{fig:framework}, the foreground layer representation $LM_f$ is encoded by a mapping module $\mathcal{M}_{f}$, a texture module $\mathcal{T}_{f}$, and a residual module $\mathcal{R}_{f}$. The mapping module $\mathcal{M}{f}$ transforms the input coordinate $p$ to the texture space coordinate $(u,v)^p_f$. The texture module $\mathcal{T}_{f}$ converts the texture coordinate into an RGB value $c_{f}$, while the residual module $\mathcal{R}_{f}$ predicts a frame-wise residual coefficient $r_{f}$. Only $\mathcal{T}{f}$ and $\mathcal{R}{f}$ incorporate Multi-resolution Hashing Encoding (MRHE) to accelerate training and inference. The MRHE (illustrated in the rightmost block of Figure \ref{fig:framework}) performs grid hash encoding~\cite{muller2022instant}. Here, the input 2D features serve as coordinates within multi-resolution grids, allowing for interpolation-based data sampling. The sampled data is then concatenated and processed through MLPs.
The final output of the foreground layer is computed as $\hat{c}^p_f=c^p_f \ast r^p_f $. The background layer follows the same structure.

The background and foreground representations are later combined by an alpha module $\mathcal{A}$. This module takes the coordinate $p$ as input and predicts a soft opacity map for each layer. The final reconstructed RGB value, $\hat{c}^p$, is obtained by weighted summation across both layers: 
\begin{equation}
    \hat{c}^p=(1-{\alpha}^p)\hat{c}^p_b+{\alpha}^p \hat{c}^p_f. 
\end{equation}
All the modules in the above model structure are MLPs. 

Note that our explanation can easily be switched to multiple foreground layers based on the masks provided by users; we explained the case of a single foreground layer for simplification.

\vspace{2mm}
\noindent \textbf{Hypernet.} 
Our hypernet, $\mathcal{H}$, predicts the INR parameters given a condition embedding $e$ ---see the Hypernet block in Figure \ref{fig:framework}. It comprises a series of MLPs, each dedicated to generating the parameters of a specific layer within the NVD model.
The target of $\mathcal{H}$ is to generate accurate weights for an NVD model capable of robust video decomposition. To accomplish this, we train it using the same loss functions and constraints applied in single-video decomposition models, as the objective for each video remains unchanged. However, by conditioning the hypernet on unique embeddings for each video, the model learns a generalized decomposition capability that extends beyond individual instances.

\vspace{2mm}
\noindent \textbf{VideoMAE encoder.} 
Typically, hypernet inputs use either learnable embeddings trained jointly with the hypernet or embeddings generated from input data. We use the latter, as representing high-dimensional video data in a compact, low-dimensional embedding is particularly challenging. This approach also supports generalization on multiple videos and enables inference on unseen videos, which is crucial for building a robust NVD model. Moreover, this video embedding needs to encapsulate both motion and scene information, help reduce optimization complexity, and accelerate network convergence. 

\begin{figure}[t]
    \centering
    \includegraphics[width=\linewidth]{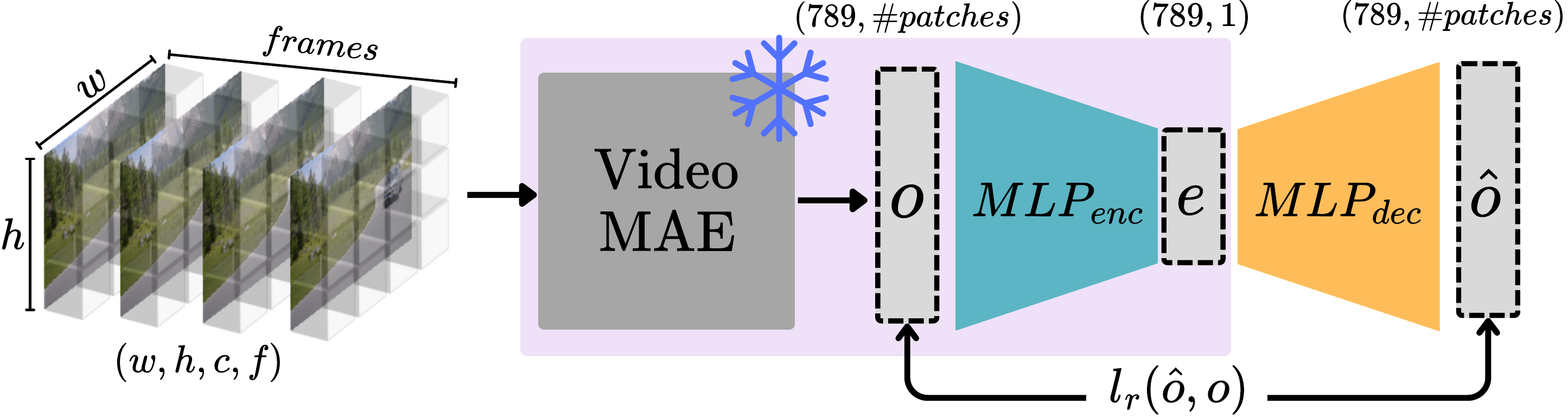}
    \caption{Training of the autoencoder to get a compressed embedding from VideoMAE. First, we extract features $o$ from each 3D patch using the pre-trained VideoMAE. We then train an autoencoder to obtain a compressed embedding $e$ by minimizing the difference between the output $\hat{o}$ and the input $o$.}
    \vspace{-2mm}
    \label{fig:mae_enc}
\end{figure}

To achieve this, we use the features predicted by VideoMAE~\cite{tong2022videomae}, a frozen self-supervised masked autoencoder trained on video data. As shown in Figure~\ref{fig:mae_enc}, to further reduce the dimension of the VideoMAE output, we use an encoder-decoder structure to learn a compressed embedding of size $(768, 1)$ from the original VideoMAE output of size $(768, \#patches)$. This ensures the encoder consolidates information from all patches into a single dense embedding while minimizing information loss. The autoencoder is trained using $L_1$ loss to enforce the closeness between the reconstructed output $\hat{o}$ and the original embedding $o$. 

We take the VideoMAE together with the encoder part as our VideoMAE encoder in HyperNVD (the top-left module in Figure~\ref{fig:framework}). This encoder is pre-trained and frozen during the hypernet training, allowing us to maintain the rich representations it learned while preserving essential information for the decomposition tasks. In this way, the encoder serves as a compact, fixed embedding provider for the hypernet.

\subsection{Training and rendering procedure}
\noindent \textbf{Loss functions.}
Since all the components of our model are fully differentiable, we train them end-to-end via stochastic gradient descent. We use losses from the previous work~\cite{kasten2021layered, chan2023hashing} to facilitate the training process. The losses include: 
\begin{itemize}
    \item \textit{Reconstruction loss} to ensure the reconstruction quality of the video.
    \item \textit{Consistency loss} which provides supervision of optical flow for accurate motion representation.
    \item \textit{Sparsity loss} to prevent duplicate content in different texture layers.
    \item \textit{Residual consistency loss} to ensure smooth lighting conditions while preventing the residual estimator from absorbing all color details.
\end{itemize}    
Two losses are applied just in the initial iterations:
    \begin{itemize}  
    \item \textit{Rigidity loss} to maintain the structural rigidity in the canonical texture layer.
    \item \textit{Alpha bootstrapping loss}, which supervises coarse masks for foreground separation. 
    \end{itemize}
    
Since the model is trained in a self-supervised manner, the additional supervision we use are the ground truth object masks and the optical flow predicted by RAFT~\cite{teed2020raft}. Further details are in the supplementary materials. 

\vspace{2mm}
\noindent \textbf{Training with the hypernet.} 
The objective is to train a hypernet capable of generating parameters for the NVD model, so only the hypernet weights are trainable, while the NVD model serves as a set of differentiable layers that enable backpropagation to the hypernet, allowing it to learn from the above-mentioned losses.
We begin the training process with a pre-training step that configures the mapping network to predict textures that are both aligned and in an initial rectangular shape corresponding to the frame size. Without this initial step, the model could produce incorrectly oriented textures. With this pre-training, the convergence during the subsequent training phase is significantly faster, as the model starts from a more informed initialization.

\vspace{2mm}
\noindent \textbf{Fine-tune on new videos.} 
Once our model is trained on a set of videos, we can use the hypernet to generate parameters for a new video and treat these weights as an initialization for the NVD model (see Figure~\ref{fig:teaser} and Section~\ref{sec:quan}). 


\vspace{2mm}
\noindent \textbf{Video editing and rendering.}
The NVD model decomposes an input video into multiple layers, each with distinct textures. For video editing, we first perform inference to obtain these texture layers. Then, each layer can be edited individually with image editing software or other techniques. For editing, we render the texture networks using a $1000 \times 1000$ grid sampling. The rendered texture is then modified, and the colors are sampled through bilinear interpolation for smooth transitions and precise adjustments. By replacing the original layers with the modified ones and mapping the corresponding coordinates back to each video frame, we can generate the edited video with temporally consistent changes.


\begin{table*}[t!]
\centering
\caption{Comparison with previous approaches on reconstruction quality for single video training.}\vspace{-1mm}
\resizebox{1.8\columnwidth}{!}{%
\setlength{\tabcolsep}{1.5mm}{
\begin{tabular}{c|cc|cc|cc|cc|cc}
\hline
\multirow{2}{*}{Methods} &
\multicolumn{2}{c|}{hike} &
\multicolumn{2}{c|}{blackswan} &
\multicolumn{2}{c|}{car-turn} &
\multicolumn{2}{c|}{bear} &
\multicolumn{2}{c}{lucia} \\
& PSNR  & SSIM  & PSNR  & SSIM  & PSNR  & SSIM  & PSNR  & SSIM  & PSNR  & SSIM  \\ \hline
Deformable Sprites~\cite{ye2022deformable}      & 25.30 & 0.795 & 25.75 & 0.732 & 30.74 & 0.850 & 26.34 & 0.816 & 26.54 & 0.771 \\
Layered Neural Atlases~\cite{kasten2021layered} & 30.02 & 0.882 & 29.96 & 0.853 & 34.39 & 0.929 & 29.62 & 0.910 & 29.92 & 0.888 \\
Hashing-nvd~\cite{chan2023hashing}              & 29.12 & 0.873 & 30.30 & 0.843 & 35.43 & 0.895 & 31.56 & 0.908 & 29.29 & 0.866 \\
\textbf{HyperNVD} (Ours)                        & 30.06 & 0.846 & 29.53 & 0.810 & 34.94 & 0.909 & 31.58 & 0.901 & 30.30 & 0.865 \\ \hline
\end{tabular}%
}
}
\label{tab:comparison}
\end{table*}

\begin{table*}[ht]
\centering
\caption{Comparison of the reconstruction quality of our model when trained with different numbers of videos.}
     \vspace{-1mm}
\resizebox{\textwidth}{!}{%
\setlength{\tabcolsep}{0.7mm}{
\begin{tabular}{c|cc|cc|cc|cc|cc|cc|cc|cc|cc|cc}
\hline
&
  \multicolumn{2}{c|}{hike} &
  \multicolumn{2}{c|}{blackswan} &
  \multicolumn{2}{c|}{car-turn} &
  \multicolumn{2}{c|}{bear} &
  \multicolumn{2}{c|}{lucia} &
  \multicolumn{2}{c|}{kite-walk} &
  \multicolumn{2}{c|}{bus} &
  \multicolumn{2}{c|}{stroller} &
  \multicolumn{2}{c|}{boat} &
  \multicolumn{2}{c}{rollerblade} \\ 
\multirow{-2}{*}{\begin{tabular}[c]{@{}c@{}}Num of \\ videos\end{tabular}} &
  PSNR &
  SSIM &
  PSNR &
  SSIM &
  PSNR &
  SSIM &
  PSNR &
  SSIM &
  PSNR &
  SSIM &
  PSNR &
  SSIM &
  PSNR &
  SSIM &
  PSNR &
  SSIM &
  PSNR &
  SSIM &
  PSNR &
  SSIM \\ \hline
1 &
  30.06 &
  0.846 &
  29.53 &
  0.810 &
  34.94 &
  0.909 &
  31.58 &
  0.901 &
  30.30 &
  0.865 &
  33.35 &
  0.837 &
  32.45 &
  0.889 &
  31.25 &
  0.830 &
  32.77 &
  0.872 &
  28.75 &
  0.810 \\
3 &
  27.31 &
  0.728 &
  27.67 &
  0.754 &
  31.87 &
  0.850 &
  - &
  - &
  - &
  - &
  - &
  - &
  - &
  - &
  - &
  - &
  - &
  - &
  - &
  - \\
5 &
  27.13 &
  0.760 &
  27.32 &
  0.734 &
  31.43 &
  0.845 &
  28.88 &
  0.854 &
  27.43 &
  0.760 &
  - &
  - &
  - &
  - &
  - &
  - &
  - &
  - &
  - &
  - \\
10 &
  27.99 &
  0.754 &
  28.86 &
  0.736 &
  32.47 &
  0.852 &
  29.99 &
  0.853 &
  28.50 &
  0.763 &
  31.75 &
  0.797 &
  29.45 &
  0.834 &
  28.80 &
  0.775 &
  31.07 &
  0.828 &
  26.06 &
  0.740 \\
15 &
  27.65 &
  0.768 &
  27.41 &
  0.756 &
  32.57 &
  0.878 &
  29.71 &
  0.852 &
  28.54 &
  0.833 &
  31.99 &
  0.796 &
  29.60 &
  0.844 &
  28.92 &
  0.780 &
  30.91 &
  0.826 &
  26.13 &
  0.742 \\
30 &
  27.24 &
   0.795 &
  27.30&
   0.745 &
  31.64&
   0.864 &
  29.28&
   0.862 &
  27.31&
   0.802 &
  30.93&
   0.771 &
  28.63&
   0.827 &
  27.00&
    0.720 &
  29.98&
   0.803 &
  24.35& 0.686  \\ \hline
\end{tabular}
}
}
\label{tab:num_video}
\end{table*}

\begin{table*}[t]
\centering
\caption{Reconstruction quality in PSNR on unseen videos when fine-tuning from our 15-video metamodel versus training from scratch.}
  \vspace{-1mm}
\label{tab:unseen_videos}
\setlength{\tabcolsep}{2mm}{
\begin{tabular}{c|cccccccccc}
\hline
Methods           & cows  & soccerball & soapbox & famingo & horse-jump  & goat  & swing & kite-surf & libby & surf  \\ \hline
From scratch      & 27.04 & 28.51      & 31.28   & 29.03    & 27.61      & 27.06 & 29.65 & 31.65     & 29.63 & 33.61 \\
From metamodel    & 28.40  & 30.16      & 32.74   & 29.96    & 28.50       & 27.78 & 30.37 & 32.19     & 30.11 & 33.86 \\
Improvements      & +1.36 & +1.65      & +1.46   & +0.93    & +0.89      & +0.72 & +0.72 & +0.54     & +0.48 & +0.25  \\ \hline
\end{tabular}%
}
\end{table*}

\section{Experiments}
\label{sec:exp}

\subsection{Implementation details}

\noindent \textbf{Datasets.} 
We conduct our experiments mainly on the DAVIS dataset~\cite{perazzi2016benchmark}. 
We train HyperNVD with up to 30 videos, while the rest videos are reserved for testing. Each video consists of the first 16 frames at a resolution of $768\times432$. 

\vspace{2mm}
\noindent \textbf{Training settings.} 
We use a batch size of 100,000 point locations and train each model for around 25,000 iterations and 100 pre-train iterations. The hypernet consists of approximately 290 million learnable parameters. The main network consists of approximately 4.4 million learnable parameters: 1.75 million for each layer module and 965 thousand for the alpha module. 
Training and evaluation with 1 video and a batch size of 100.000 requires around 22GB of GPU memory, with a checkpoint size of 2.3GB. Training our model takes roughly 2.8 hours for 1 video and 12.5 hours for 3 videos on an NVIDIA RTX 3090 GPU. We optimize all networks simultaneously using the Adam optimizer, with a learning rate of 1e-2 for the hash encoding and a learning rate of 5e-4 for the others.

\vspace{2mm}
\noindent \textbf{Details of architecture.} 
In the target NVD model, we apply Rectified Linear Unit (ReLU) activations between hidden layers, with outputs passed through a tanh function. This enables the model to map texture to a predefined continuous range.
The hypernet comprises a series of MLPs. Each MLP consists of four fully connected layers, with a hidden dimension of 64 and intermediate ReLU activations. Further details about the architecture can be found in the supplementary material.

\subsection{Quantitative results}
\label{sec:quan}

\noindent \textbf{Comparison with previous methods.} 
We present the comparison of PSNR and SSIM values between Deformable Sprites~\cite{ye2022deformable}, Layered Neural Atlases~\cite{kasten2021layered}, Hashing-nvd~\cite{chan2023hashing} and our HyperNVD in Table~\ref{tab:comparison}. We report the results of our model trained on one single video independently. For a fair comparison, we use the ground truth mask as initial guidance for all the comparing methods. Results in Table~\ref{tab:comparison} demonstrate that incorporating the hypernet does not degrade the model's performance. Ours maintains the effectiveness of the original architecture (\ie~our baseline Hashing-nvd) while introducing the flexibility of a hypernet-based approach.

\vspace{2mm}
\noindent \textbf{Learning multiple videos.}
To evaluate the model’s generalization ability across multiple videos, we train HyperNVD with 1, 3, 5, 10, 15, and 30 videos, respectively. For consistency, videos included in smaller-scale settings are subsets of those in larger-scale settings. Table~\ref{tab:num_video} shows the PSNR of the videos reconstructed by models trained with different numbers of videos. When training with multiple videos, we observe a slight drop in reconstruction quality compared to training on a single video. This is likely due to the model potentially overfitting to the unique characteristics of a single video. However, the PSNR only decreases by approximately 3dB as the number of videos increases from 1 to 30, which indicates that the model retains strong performance even with larger datasets.

\begin{figure}[t]
  \centering
   \includegraphics[width=\linewidth]{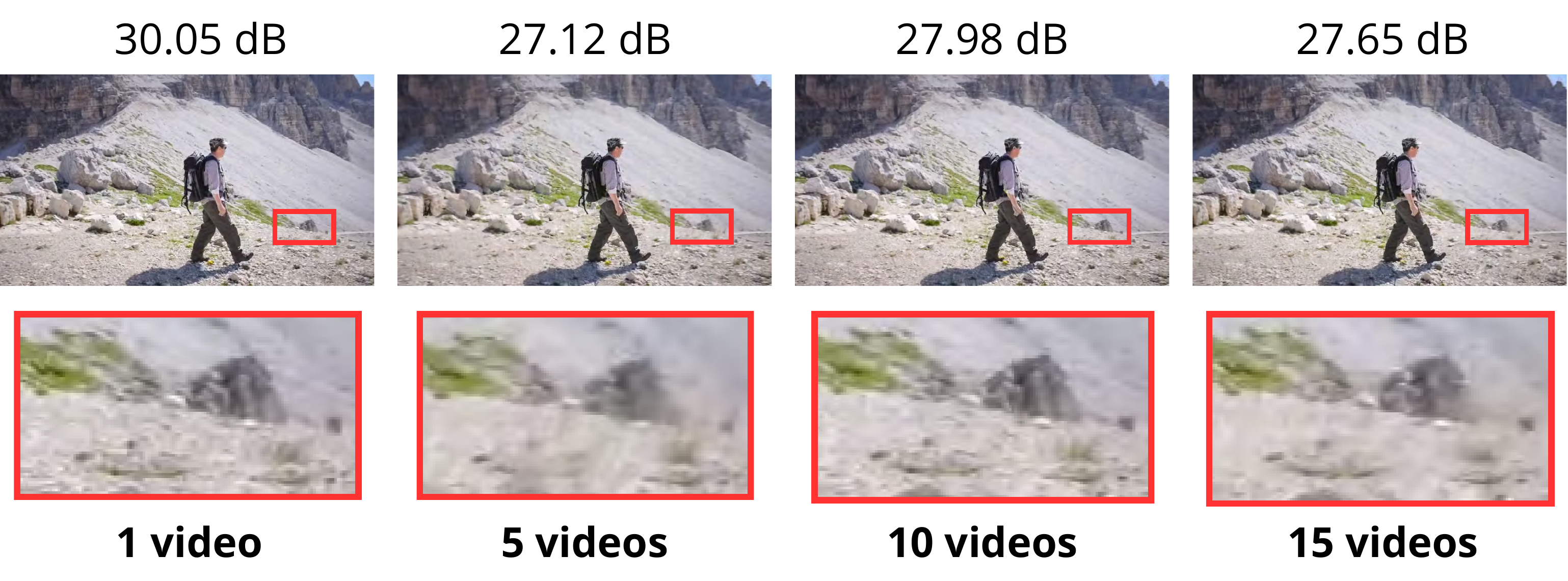}
    \vspace{-5mm}
  \caption{Comparison of the video ``hike'' reconstructed by models trained with different numbers of videos. The top row shows the full frames, while the bottom row zooms in on a specific patch. Models trained on multiple videos capture the primary structure of the frame but tend to lose some high-frequency details.}
  \label{fig:hike}
\end{figure}

\begin{figure}[t]
  \centering
   \includegraphics[width=\linewidth]{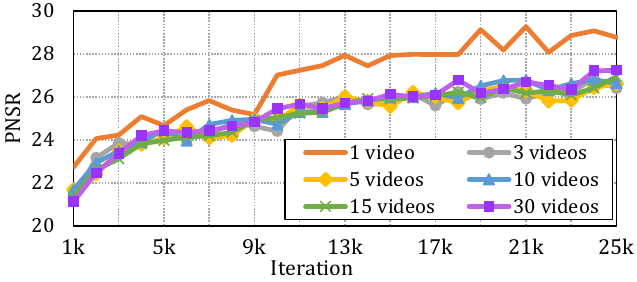}
    \vspace{-6mm}
   \caption{Comparison of the reconstruction quality for the ``hike'' video trained with different numbers of videos. Our HyperNVD model trained on multiple videos shows a 3dB drop in PSNR compared to single-video training. The PSNR remains stable across multiple video training, suggesting a performance plateau.}
   \vspace{-1mm}
   \label{fig:num_video}
\end{figure}

Figure~\ref{fig:hike} further illustrates the image quality of the ``hike'' video when training HyperNVD with different numbers of videos. The models trained on multiple videos can still reconstruct the low-frequency parts well, but some high-frequency details are lost. This stable evolution of PSNR indifferent to the number of videos indicates a plateau in performance with larger video sets. Figure~\ref{fig:num_video} shows the reconstruction quality of the video ``hike'' across different iteration steps. The resulting curves are very similar for all the cases trained with multiple videos.

\vspace{2mm}
\noindent \textbf{Finetuning on new videos.}
To evaluate the ability of fast adaption on unseen videos, we train the NVD model initialized by the parameters of our metamodel trained with 15 videos and compare it with the model trained from scratch. Note that the NVD model is trained independently without the hypernet. As shown in Table~\ref{tab:unseen_videos}, training from our metamodel (\ie~HyperNVD) consistently improves performance by 0.8 dB across different videos compared to training from scratch (\ie~Hashing-nvd). Figure~\ref{fig:teaser} further shows our model reaches the same PSNR on the unseen videos more than 30 minutes faster, and with more stable traning. This demonstrates that our metamodel enables faster adaptation to new scenes, which improves the efficiency of video editing workflow. Additional results are provided in the supplementary material.

\subsection{Qualitative results}

\begin{figure*}[ht]
    \centering
    \includegraphics[width=1\linewidth]{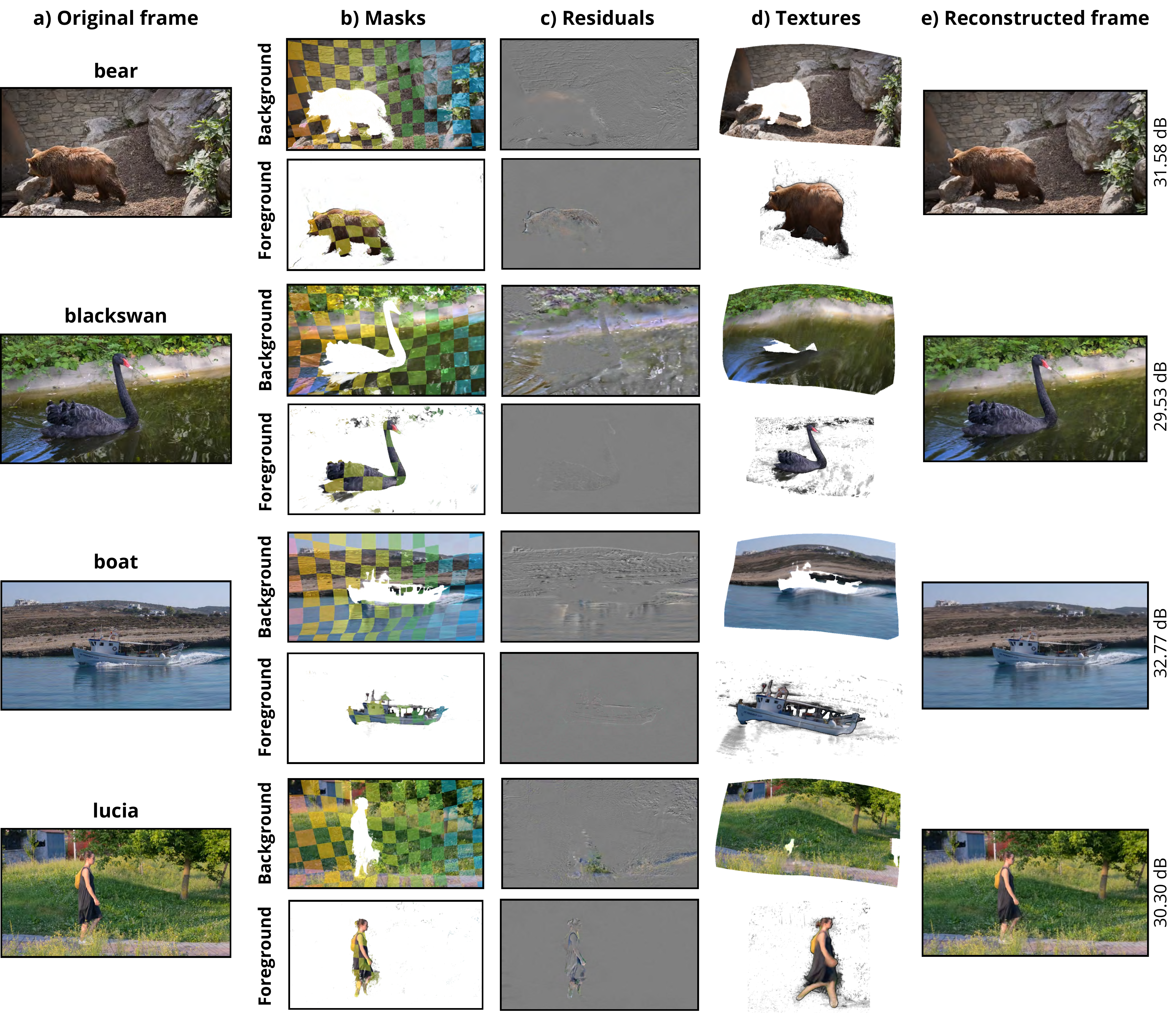}
    \caption{Qualitative results on the DAVIS dataset. Displayed are the a) original frames, b) predicted masks, c) residual maps, d) texture maps, e) reconstructed frames, and PSNR values for different videos. A color checkerboard overlay is applied to the masked areas to visualize texture transformations.}
    \label{fig:decompose}
    \vspace{-2.5mm}
\end{figure*}

Figure~\ref{fig:decompose} shows examples of our layer decomposition results. For each video frame, we include a) the original frame, b) the predicted background and foreground components with a checkerboard overlay to illustrate texture mapping transformations, c) the predicted residual maps, d) the complete canonical texture map, and e) the reconstructed frame. Our model obtains accurate video reconstruction without introducing noticeable texture distortions. The predicted opacity map separates the foreground and background correctly. The residuals learn the changing reflections on the water and shadows as shown in the ``blackswan'' and ``boat'' videos.

\begin{figure*}[t]
    \centering
    \includegraphics[width=1\linewidth]{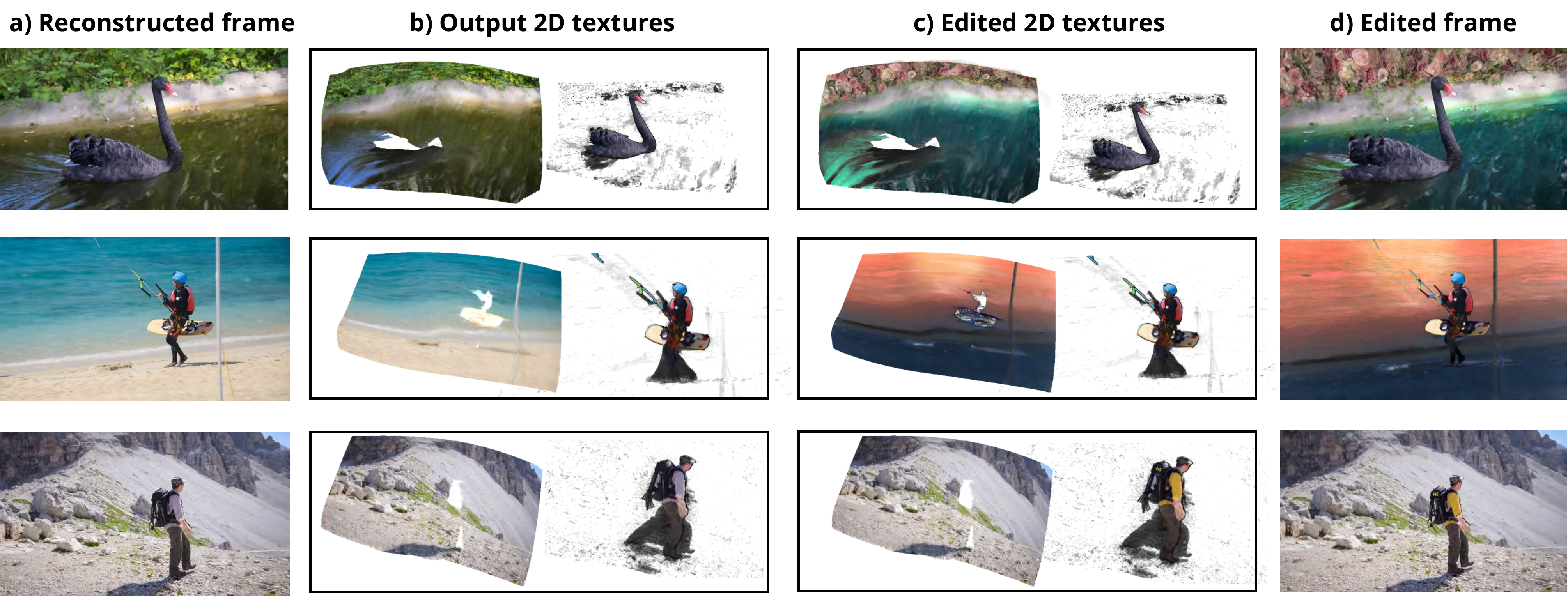}
    \caption{Edited video results. We make adjustments to one of the texture layers ---the background layers (rows 1 and 2), and the foreground layer (row 3)--- while keeping the other layer unchanged and blending the edited layers together. The edited video maintains good visual quality and smooth transitions.}
    \label{fig:edit}
  \vspace{-2mm}
\end{figure*}

To further showcase the effectiveness of our approach for editing, we demonstrate editing examples in Figure~\ref{fig:edit}. In the first two examples, only the background is modified, while in the last example, we edit the foreground texture. For the first example, the ``blackswan'' video, we transform the plants in the background into flowers and adjust the color of the water. The ripples and shadows in the water maintain a natural appearance, and the transitions at the boundaries between the swan and the water are seamless. In the second video (``kite-walk''), we alter the beach background to a sunset style. In the ``hike'' video, we change the color of the person’s shirt to yellow and add text on the backpack. Additional results of decomposition and editing are available in the supplementary material.

\subsection{Ablation study}

We do ablations on using different embedding as the input of the model, on learning the multi-resolution hash encoding with and without hypernet, and on using different types of masks as the initial supervision. 

\begin{table}[t]
\centering
\caption{Ablations on the input video embedding and learning the MRHE with or without hypernet. `Num' represents the number of videos used for training. Using the MAE encoder generated embedding is better than a learnable embedding jointly optimized with the hypernet. The video-specific MRHE encodings predicted by the hypernet outperform a universal MRHE encoding.}
\resizebox{\columnwidth}{!}{%
\begin{tabular}{c|cc|cc|cc|cc}
\hline
 & Embed & MRHE & \multicolumn{2}{c|}{hike} & \multicolumn{2}{c|}{car-turn} & \multicolumn{2}{c}{blackswan} \\
\multirow{-2}{*}{Num} & w MAE & w hyper & PSNR & SSIM & PSNR & SSIM & PSNR & SSIM \\ \hline
 & \xmark & \checkmark & 30.17 & 0.867 & 34.04 & 0.905 & 29.95 & 0.822 \\
 & \checkmark & \xmark & 30.23 & 0.858 & 33.47 & 0.881 & 28.83 & 0.797 \\
\multirow{-3}{*}{1} & \textbf{\checkmark} & \textbf{\checkmark} & 30.06 & 0.846 & 34.94 & 0.909 & 29.53 & 0.810 \\ \hline
 & \xmark & \checkmark & 26.37 & 0.760 & 29.98 & 0.829 & 26.07 & 0.684 \\
 & \checkmark & \xmark & 26.88 & 0.712 & 30.01 & 0.788 & 26.64 & 0.707 \\
\multirow{-3}{*}{3} & \textbf{\checkmark} & \textbf{\checkmark} & 27.31 & 0.728 & 31.87 & 0.850 & 27.67 & 0.754 \\ \hline
\end{tabular}%
}
\label{tab:ablations}
\end{table}


\vspace{2mm}
\noindent \textbf{VideoMAE embedding \textit{v.s.}~Learnable embedding.}
A well-designed embedding is crucial for a hypernet to generate effective model parameters. Some related works~\cite{chan2023hashing, maiya2024latent} employ a learnable embedding optimized jointly with the hypernet, while we directly grab the embedding from the pre-trained MAE encoder and freeze it during hypernet training. We compare the results of using an optimized embedding versus an embedding generated by the MAE encoder, as shown in the first and third rows of Table~\ref{tab:ablations}. The advantage in PSNR for the VideoMAE embedding when training for three videos is more than 1 dB for the 3-video model. For single-video training, the impact of using either embedding type on the final video quality is minimal. Thus, the advantage of using the VideoMAE Encoder becomes more evident when applied to multiple videos. VideoMAE embedding also shows superiority in unseen videos when using the 3-video metamodel as initialization (see the supplementary materials). Furthermore, using VideoMAE enables robust generalization to new, unseen videos, which is essential for the versatility of our model.

\vspace{2mm}
\noindent \textbf{Video-based MRHE learned with hypernet \textit{v.s.} Universal MRHE without hypernet.} 
In this part, we show the benefits of learning multiresolution hash encoding with the hypernet. In multiresolution hash encoding, input 2D features are treated as coordinates within multi-resolution grids and used to sample data from these grids via interpolation. The sampled data are then concatenated and processed through MLPs to generate the final output. In Table~\ref{tab:ablations}, we compare the results of i) using hypernet to learn the parameters of the MLPs making these parameters unique for each video (third rows) and ii) directly using the main NVD net to learn MRHE parameters making them common for all videos (second rows). The higher quality results are obtained when MRHE is learned by the hypernet, with benefits that become more apparent as the number of videos used for training increases. This improvement stems from the fact that different video scenes exhibit varying levels of complexity, making video-specific hash grid parameters beneficial when training across multiple videos.

\begin{figure}[t]
  \centering
   \includegraphics[width=\linewidth]{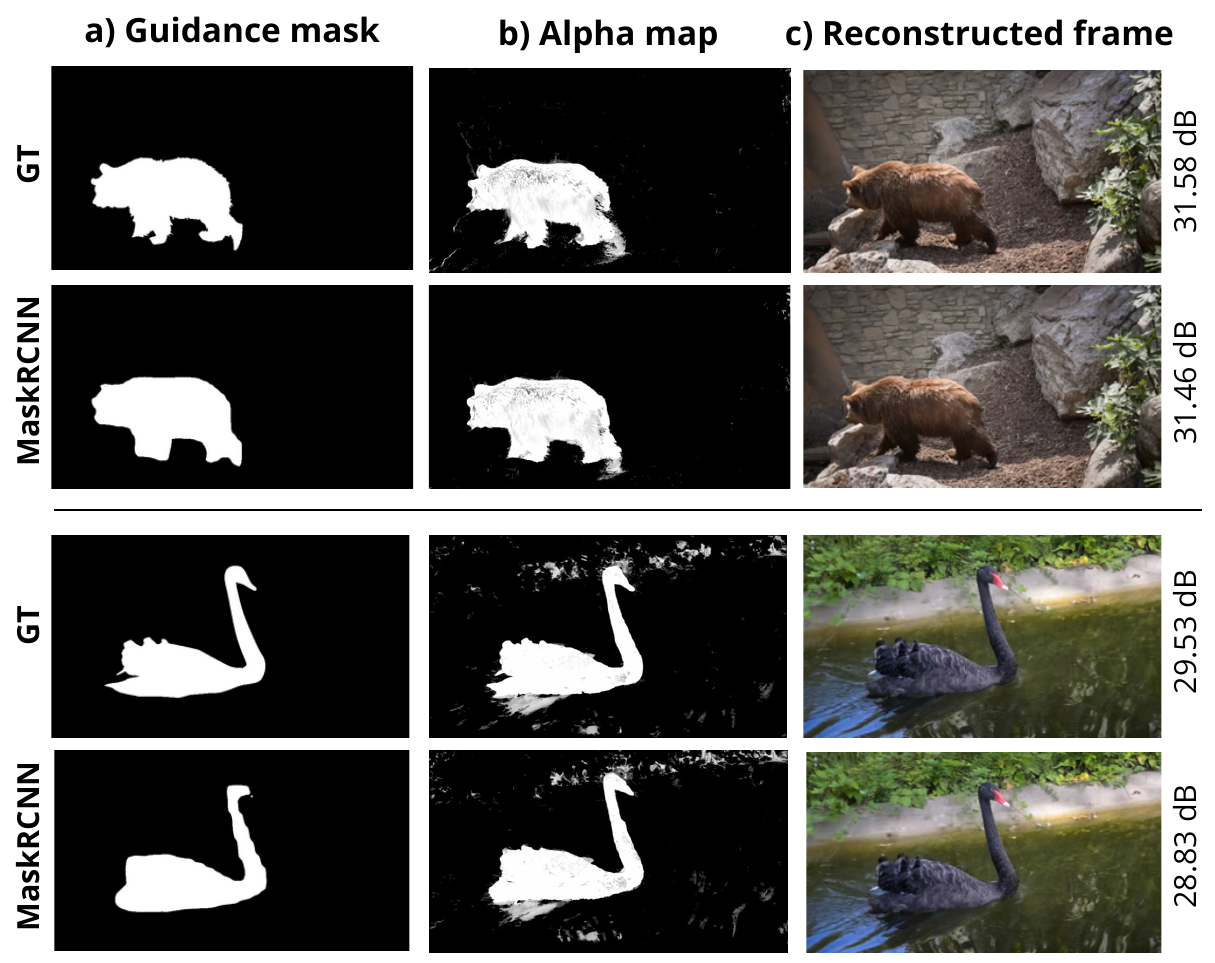}
  \caption{Comparison between using accurate and coarse masks as initial guidance. Using different masks does not significantly affect the reconstruction results, but providing a refined mask as the initial guidance helps the model achieve more detailed decomposition, which can be especially beneficial for finer edits.}
  \label{fig:mask}
   \vspace{-2mm}
\end{figure}

\vspace{2mm}
\noindent \textbf{Accurate mask \textit{v.s.} Coarse mask.}
The reference mask is important in our task since it provides a clue about the moving salient objects. We compare the use of manually annotated precise masks (groundtruth from the DAVIS dataset~\cite{perazzi2016benchmark}) with coarsely predicted masks from MaskRCNN~\cite{he2017mask} as the initial reference for model training. The predicted alpha map is a soft mask that always includes the corresponding effects, such as reflection or shadows accompanied by the moving objects. Therefore, it is always different from the original object mask.
Results in Table~\ref{tab:ablation_mask} show that our model is robust to masks with varying levels of accuracy. Figure~\ref{fig:mask} illustrates two frames from the ``bear'' and ``blackswan'' videos, where a more precise initial mask improves object boundary details. For example, using a more refined mask as initial guidance help the alpha map to capture intricate structures, such as the feathers on a swan's tail. Thus, for higher-quality editing results, it is essential to use the most detailed masks possible during model training.

\begin{table}[t]
\centering
\caption{Ablation on using different masks as initial guidance.}
 \vspace{-1mm}
\centering
\resizebox{0.98\linewidth}{!}{%
\setlength{\tabcolsep}{1.0mm}{
\begin{tabular}{c|cc|cc|cc}
\hline
\multirow{2}{*}{Mask} & \multicolumn{2}{c|}{hike}              & \multicolumn{2}{c|}{car-turn}          & \multicolumn{2}{c}{blackswan}          \\
                      & PSNR  & SSIM   & PSNR  & SSIM   & PSNR  & SSIM   \\ \hline
GT                    & 30.06 & 0.846 & 34.94 & 0.909 & 29.53 & 0.810 \\
MaskRCNN              & 29.61 & 0.846 & 34.92 & 0.903 & 29.66 & 0.814 \\ \hline
\end{tabular}%
}
}
\label{tab:ablation_mask}
\end{table}

\section{Concluding Remarks}

We present a framework for multiple video training of INR-based video decomposition models using a hypernetwork. We reformulate the task of training a neural video decomposition model on a single video into learning a hypernetwork that generates network parameters based on video embeddings. This approach effectively mitigates the overfitting of the neural video decomposition model to individual videos, enhances generalization capability, and speeds up adaptation to unseen videos. Through experiments, we validate that our proposed model achieves comparable video layer decomposition and rendering capabilities to previous methods, while significantly reducing training time on new videos and improving video editing efficiency.

\vspace{3mm}
{
\noindent \textbf{Acknowledgement.} 
We acknowledge the Departament de Recerca i Universitats with ref. 2021SGR01499 and the CERCA Program from Generalitat de Catalunya, Grant PID2021-128178OB-I00 funded by MCIN/AEI/10.13039/501100011033, ERDF "A way of making Europe", and the grant Càtedra ENIA UAB-Cruïlla (TSI-100929-2023-2) from the Ministry of Economic Affairs and Digital Transformation of Spain.
}

\newpage

{
    \small
    \bibliographystyle{ieeenat_fullname}
    \bibliography{main}

\begin{thebibliography}{41}
\providecommand{\natexlab}[1]{#1}
\providecommand{\url}[1]{\texttt{#1}}
\expandafter\ifx\csname urlstyle\endcsname\relax
  \providecommand{\doi}[1]{doi: #1}\else
  \providecommand{\doi}{doi: \begingroup \urlstyle{rm}\Url}\fi

\bibitem[Adobe()]{AdobeAE}
Adobe.
\newblock After effects.
\newblock Accessed: 2024-11-07.

\bibitem[Athar et~al.(2023)Athar, Hermans, Luiten, Ramanan, and Leibe]{athar2023tarvis}
Ali Athar, Alexander Hermans, Jonathon Luiten, Deva Ramanan, and Bastian Leibe.
\newblock Tarvis: A unified approach for target-based video segmentation.
\newblock In \emph{CVPR}, 2023.

\bibitem[BlackMagicDesign()]{DaVinci}
BlackMagicDesign.
\newblock Davinci resolve 19.
\newblock Accessed: 2024-11-07.

\bibitem[Chan et~al.(2023)Chan, Yuan, Sun, and Chen]{chan2023hashing}
Cheng-Hung Chan, Cheng-Yang Yuan, Cheng Sun, and Hwann-Tzong Chen.
\newblock Hashing neural video decomposition with multiplicative residuals in space-time.
\newblock In \emph{ICCV}, 2023.

\bibitem[Chauhan et~al.(2024)Chauhan, Zhou, Lu, Molaei, and Clifton]{chauhan2024brief}
Vinod~Kumar Chauhan, Jiandong Zhou, Ping Lu, Soheila Molaei, and David~A Clifton.
\newblock A brief review of hypernetworks in deep learning.
\newblock \emph{Artificial Intelligence Review}, 57\penalty0 (9):\penalty0 250, 2024.

\bibitem[Chiang et~al.(2022)Chiang, Tsai, Tseng, Lai, and Chiu]{chiang2022stylizing}
Pei-Ze Chiang, Meng-Shiun Tsai, Hung-Yu Tseng, Wei-Sheng Lai, and Wei-Chen Chiu.
\newblock Stylizing 3d scene via implicit representation and hypernetwork.
\newblock In \emph{WACV}, 2022.

\bibitem[Cong et~al.(2023)Cong, Xu, Simon, Chen, Ren, Xie, Perez-Rua, Rosenhahn, Xiang, and He]{cong2023flatten}
Yuren Cong, Mengmeng Xu, Christian Simon, Shoufa Chen, Jiawei Ren, Yanping Xie, Juan-Manuel Perez-Rua, Bodo Rosenhahn, Tao Xiang, and Sen He.
\newblock Flatten: optical flow-guided attention for consistent text-to-video editing.
\newblock In \emph{ICLR}, 2023.

\bibitem[Doersch et~al.(2022)Doersch, Gupta, Markeeva, Recasens, Smaira, Aytar, Carreira, Zisserman, and Yang]{doersch2022tap}
Carl Doersch, Ankush Gupta, Larisa Markeeva, Adria Recasens, Lucas Smaira, Yusuf Aytar, Joao Carreira, Andrew Zisserman, and Yi Yang.
\newblock Tap-vid: A benchmark for tracking any point in a video.
\newblock \emph{NeurIPS}, 2022.

\bibitem[Finn et~al.(2017)Finn, Abbeel, and Levine]{finn2017model}
Chelsea Finn, Pieter Abbeel, and Sergey Levine.
\newblock Model-agnostic meta-learning for fast adaptation of deep networks.
\newblock In \emph{ICML}, 2017.

\bibitem[Foundry()]{nuke}
Foundry.
\newblock Nuke.
\newblock Accessed: 2024-11-07.

\bibitem[Geyer et~al.(2024)Geyer, Bar-Tal, Bagon, and Dekel]{geyer2023tokenflow}
Michal Geyer, Omer Bar-Tal, Shai Bagon, and Tali Dekel.
\newblock Tokenflow: Consistent diffusion features for consistent video editing.
\newblock In \emph{ICLR}, 2024.

\bibitem[Gu et~al.(2023)Gu, Xian, Snavely, and Davis]{gu2023factormatte}
Zeqi Gu, Wenqi Xian, Noah Snavely, and Abe Davis.
\newblock Factormatte: Redefining video matting for re-composition tasks.
\newblock \emph{ACM TOG}, 42\penalty0 (4):\penalty0 1--14, 2023.

\bibitem[Ha et~al.(2017)Ha, Dai, and Le]{ha2016hypernetworks}
David Ha, Andrew Dai, and Quoc~V Le.
\newblock Hypernetworks.
\newblock In \emph{ICLR}, 2017.

\bibitem[He et~al.(2017)He, Gkioxari, Doll{\'a}r, and Girshick]{he2017mask}
Kaiming He, Georgia Gkioxari, Piotr Doll{\'a}r, and Ross Girshick.
\newblock Mask r-cnn.
\newblock In \emph{ICCV}, 2017.

\bibitem[Jojic and Frey(2001)]{jojic2001learning}
Nebojsa Jojic and Brendan~J Frey.
\newblock Learning flexible sprites in video layers.
\newblock In \emph{CVPR}, 2001.

\bibitem[Karaev et~al.(2024)Karaev, Rocco, Graham, Neverova, Vedaldi, and Rupprecht]{karaev2023cotracker}
Nikita Karaev, Ignacio Rocco, Benjamin Graham, Natalia Neverova, Andrea Vedaldi, and Christian Rupprecht.
\newblock Cotracker: It is better to track together.
\newblock In \emph{ECCV}, 2024.

\bibitem[Kasten et~al.(2021)Kasten, Ofri, Wang, and Dekel]{kasten2021layered}
Yoni Kasten, Dolev Ofri, Oliver Wang, and Tali Dekel.
\newblock Layered neural atlases for consistent video editing.
\newblock \emph{ACM TOG}, 40\penalty0 (6):\penalty0 1--12, 2021.

\bibitem[KeenTools()]{GeoTracker}
KeenTools.
\newblock Geotracker for adobe after effects.
\newblock Accessed: 2024-11-07.

\bibitem[Kim et~al.(2023)Kim, Lee, Kim, Cho, and Han]{kim2023generalizable}
Chiheon Kim, Doyup Lee, Saehoon Kim, Minsu Cho, and Wook-Shin Han.
\newblock Generalizable implicit neural representations via instance pattern composers.
\newblock In \emph{CVPR}, 2023.

\bibitem[Lee et~al.(2024)Lee, Kim, and Cho]{lee2024rna}
Jaekyeong Lee, Geonung Kim, and Sunghyun Cho.
\newblock Rna: Video editing with roi-based neural atlas.
\newblock \emph{ACCV}, 2024.

\bibitem[Lu et~al.(2021)Lu, Cole, Dekel, Zisserman, Freeman, and Rubinstein]{lu2021omnimatte}
Erika Lu, Forrester Cole, Tali Dekel, Andrew Zisserman, William~T Freeman, and Michael Rubinstein.
\newblock Omnimatte: Associating objects and their effects in video.
\newblock In \emph{CVPR}, 2021.

\bibitem[Maiya et~al.(2024)Maiya, Gupta, Gwilliam, Ehrlich, and Shrivastava]{maiya2024latent}
Shishira~R Maiya, Anubhav Gupta, Matthew Gwilliam, Max Ehrlich, and Abhinav Shrivastava.
\newblock Latent-inr: A flexible framework for implicit representations of videos with discriminative semantics.
\newblock In \emph{ECCV}, 2024.

\bibitem[M{\"u}ller et~al.(2022)M{\"u}ller, Evans, Schied, and Keller]{muller2022instant}
Thomas M{\"u}ller, Alex Evans, Christoph Schied, and Alexander Keller.
\newblock Instant neural graphics primitives with a multiresolution hash encoding.
\newblock \emph{ACM TOG}, 41\penalty0 (4):\penalty0 1--15, 2022.

\bibitem[Nam et~al.(2022)Nam, Brubaker, and Brown]{nam2022neural}
Seonghyeon Nam, Marcus~A Brubaker, and Michael~S Brown.
\newblock Neural image representations for multi-image fusion and layer separation.
\newblock In \emph{ECCV}, 2022.

\bibitem[Nichol et~al.(2018)Nichol, Achiam, and Schulman]{Nichol2018OnFM}
Alex Nichol, Joshua Achiam, and John Schulman.
\newblock On first-order meta-learning algorithms.
\newblock \emph{arXiv preprint arXiv:1803.02999}, 2018.

\bibitem[Ouyang et~al.(2024)Ouyang, Wang, Xiao, Bai, Zhang, Zheng, Zhou, Chen, and Shen]{ouyang2024codef}
Hao Ouyang, Qiuyu Wang, Yuxi Xiao, Qingyan Bai, Juntao Zhang, Kecheng Zheng, Xiaowei Zhou, Qifeng Chen, and Yujun Shen.
\newblock Codef: Content deformation fields for temporally consistent video processing.
\newblock In \emph{CVPR}, 2024.

\bibitem[Perazzi et~al.(2016)Perazzi, Pont-Tuset, McWilliams, Van~Gool, Gross, and Sorkine-Hornung]{perazzi2016benchmark}
Federico Perazzi, Jordi Pont-Tuset, Brian McWilliams, Luc Van~Gool, Markus Gross, and Alexander Sorkine-Hornung.
\newblock A benchmark dataset and evaluation methodology for video object segmentation.
\newblock In \emph{CVPR}, 2016.

\bibitem[Rav-Acha et~al.(2008)Rav-Acha, Kohli, Rother, and Fitzgibbon]{rav2008unwrap}
Alex Rav-Acha, Pushmeet Kohli, Carsten Rother, and Andrew Fitzgibbon.
\newblock Unwrap mosaics: A new representation for video editing.
\newblock In \emph{SIGGRAPH}, 2008.

\bibitem[Ravi et~al.(2024)Ravi, Gabeur, Hu, Hu, Ryali, Ma, Khedr, R{\"a}dle, Rolland, Gustafson, et~al.]{ravi2024sam}
Nikhila Ravi, Valentin Gabeur, Yuan-Ting Hu, Ronghang Hu, Chaitanya Ryali, Tengyu Ma, Haitham Khedr, Roman R{\"a}dle, Chloe Rolland, Laura Gustafson, et~al.
\newblock Sam 2: Segment anything in images and videos.
\newblock \emph{arXiv preprint arXiv:2408.00714}, 2024.

\bibitem[Sen et~al.(2024)Sen, Singh, Agarwal, Agaram, Krishna, and Sridhar]{sen2024hyp}
Bipasha Sen, Gaurav Singh, Aditya Agarwal, Rohith Agaram, Madhava Krishna, and Srinath Sridhar.
\newblock Hyp-nerf: Learning improved nerf priors using a hypernetwork.
\newblock \emph{NeurIPS}, 2024.

\bibitem[Sengupta et~al.(2020)Sengupta, Jayaram, Curless, Seitz, and Kemelmacher-Shlizerman]{sengupta2020background}
Soumyadip Sengupta, Vivek Jayaram, Brian Curless, Steven~M Seitz, and Ira Kemelmacher-Shlizerman.
\newblock Background matting: The world is your green screen.
\newblock In \emph{CVPR}, 2020.

\bibitem[Skorokhodov et~al.(2021)Skorokhodov, Ignatyev, and Elhoseiny]{skorokhodov2021adversarial}
Ivan Skorokhodov, Savva Ignatyev, and Mohamed Elhoseiny.
\newblock Adversarial generation of continuous images.
\newblock In \emph{CVPR}, 2021.

\bibitem[Smirnov et~al.(2021)Smirnov, Gharbi, Fisher, Guizilini, Efros, and Solomon]{smirnov2021marionette}
Dmitriy Smirnov, Michael Gharbi, Matthew Fisher, Vitor Guizilini, Alexei Efros, and Justin~M Solomon.
\newblock Marionette: Self-supervised sprite learning.
\newblock \emph{NeurIPS}, 2021.

\bibitem[Su et~al.(2020)Su, Yan, Zhu, Zhang, Ge, Sun, and Zhang]{su2020blindly}
Shaolin Su, Qingsen Yan, Yu Zhu, Cheng Zhang, Xin Ge, Jinqiu Sun, and Yanning Zhang.
\newblock Blindly assess image quality in the wild guided by a self-adaptive hyper network.
\newblock In \emph{CVPR}, 2020.

\bibitem[Teed and Deng(2020)]{teed2020raft}
Zachary Teed and Jia Deng.
\newblock Raft: Recurrent all-pairs field transforms for optical flow.
\newblock In \emph{ECCV}, 2020.

\bibitem[Tong et~al.(2022)Tong, Song, Wang, and Wang]{tong2022videomae}
Zhan Tong, Yibing Song, Jue Wang, and Limin Wang.
\newblock Video{MAE}: Masked autoencoders are data-efficient learners for self-supervised video pre-training.
\newblock In \emph{NeurIPS}, 2022.

\bibitem[von Oswald et~al.(2020)von Oswald, Henning, Grewe, and Sacramento]{Oswald2020Continual}
Johannes von Oswald, Christian Henning, Benjamin~F. Grewe, and João Sacramento.
\newblock Continual learning with hypernetworks.
\newblock In \emph{ICLR}, 2020.

\bibitem[Wang and Adelson(1994)]{wang1994representing}
John~YA Wang and Edward~H Adelson.
\newblock Representing moving images with layers.
\newblock \emph{IEEE TIP}, 3\penalty0 (5):\penalty0 625--638, 1994.

\bibitem[Xu et~al.(2017)Xu, Price, Cohen, and Huang]{xu2017deep}
Ning Xu, Brian Price, Scott Cohen, and Thomas Huang.
\newblock Deep image matting.
\newblock In \emph{CVPR}, 2017.

\bibitem[Ye et~al.(2022)Ye, Li, Tucker, Kanazawa, and Snavely]{ye2022deformable}
Vickie Ye, Zhengqi Li, Richard Tucker, Angjoo Kanazawa, and Noah Snavely.
\newblock Deformable sprites for unsupervised video decomposition.
\newblock In \emph{CVPR}, 2022.

\bibitem[Yu et~al.(2023)Yu, Blackburn-Matzen, Nguyen, Wang, Habib~Kazi, and Bousseau]{yu2023videodoodles}
Emilie Yu, Kevin Blackburn-Matzen, Cuong Nguyen, Oliver Wang, Rubaiat Habib~Kazi, and Adrien Bousseau.
\newblock Videodoodles: Hand-drawn animations on videos with scene-aware canvases.
\newblock \emph{ACM TOG}, 42\penalty0 (4):\penalty0 1--12, 2023.

\end{thebibliography}
}

\clearpage

\setcounter{page}{1}
\maketitlesupplementary

\renewcommand{\thesection}{\Alph{section}}    
\renewcommand{\thesubsection}{\thesection.\arabic{subsection}}  
\setcounter{section}{0}

In this supplementary material, we present more quantitative results on unseen videos in Section~\ref{sec: results}, highlighting our model's superior adaptability and efficiency for new videos. More ablation results about using the MAE encoder embedding are presented in Section~\ref{sec: ablation}. We also explain the network architectures, loss functions, and training data we use in Section~\ref{sec: details}. Finally, we showcase additional qualitative results demonstrating video layer decomposition and editing capabilities in Section~\ref{sec: quality}. Please refer to our project page \href{https://hypernvd.github.io/}{\textcolor{Bittersweet}{https://hypernvd.github.io/}} to view more video decomposition and editing results.

\section{More results on unseen videos}
\label{sec: results}

We present more results of fine-tuning the NVD model from the 15-video metamodel on unseen videos. In this setting, the parameters generated by the MAE encoder and hypernet are used to initialize the NVD model. This latter model is updated during fine-tuning. In Table~\ref{tab:unseen_wohyper_sup}, we compare the results of training the NVD model from metamodel (\ie~our HyperNVD) versus training from scratch (\ie~Hashing-nvd~\cite{chan2023hashing}). All the models are trained for 80k iterations, each video takes around 2 hours. The results show consistent improvements on all the unseen videos.

\begin{table*}[h]
\centering
\caption{Reconstruction quality (PSNR) on unseen videos when fine-tuning from our 15-video metamodel versus training from scratch.}
\label{tab:unseen_wohyper_sup}
\setlength{\tabcolsep}{1.1mm}{
\begin{tabular}{c|cccccccc}
\hline
Methods & scooter-black & car-shadow & bmx-trees & scooter-gray & mallard-water & drift-straight & rihno & paragliding \\ \hline
From scratch      & 26.12 & 30.76 & 29.20 & 30.68 & 29.12 & 29.50 & 32.66 & 32.54 \\
From metamodel    & 28.52 & 31.74 & 30.11 & 31.57 & 29.99 & 30.24 & 32.97 & 32.96 \\
Improvements      & +2.4  & +0.98 & +0.91 & +0.89 & +0.87 & +0.74 & +0.31 & +0.42 \\ \hline
\end{tabular}%
}
\end{table*}

\begin{figure*}[h]
\includegraphics[width=1\linewidth]{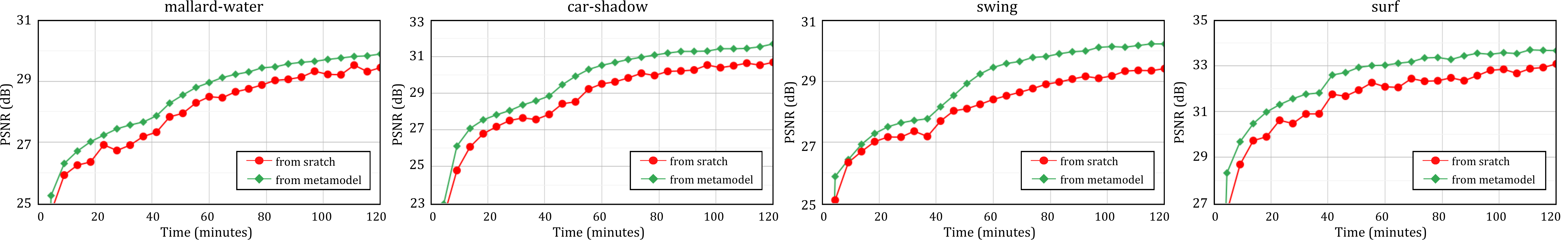}
\caption{Fine-tuning the video decomposition model from our 15-video metamodel (HyperNVD) versus training from scratch (Hashing-nvd) on unseen videos.}
\label{fig:curves_sup}
\end{figure*}

To demonstrate the acceleration of our method, the training curve of different videos is shown in Figure~\ref{fig:curves_sup}.  Our model gets the same PSNR as Hashing-nvd~\cite{chan2023hashing} (the current SOTA both for efficiency and quality, also our baseline) in 40 minutes faster on an NVIDIA RTX 3090 GPU, reflecting a 30\% improvement in training time (2 hours in total). 

\section{More ablation results}
\label{sec: ablation}
To further validate the generalization ability of the VideoMAE embeddings, we conduct an ablation on fine-tuning from the metamodel trained with 3 videos with different embeddings. Table~\ref{tab:ablations_unseen} shows that VideoMAE still outperforms learned embeddings on unseen videos. Notably, VideoMAE Encoder does not add extra computational cost during fine-tuning, so it also improves the effciency when adapting the HyperNVD to unseen videos.

\begin{table}[h]
\centering
\caption{Ablations on different embeddings on unseen videos. Results are reported in PSNR.}

\label{tab:ablations_unseen}
\begin{tabular}{c|ccc}
\hline
Embedding         &boat &bear &kite-walk \\ \hline
learned           &32.62 &30.98 &31.88 \\ 
MAE encoder       &33.31 &32.45 &32.69 \\ \hline
\end{tabular}%
\end{table}
\section{Implementation details}
\label{sec: details}

In this section, we present the detailed network architectures in Section~\ref{sec:archi}, our loss functions for training the network in Section~\ref{sec:loss}, and the videos we use for training the HyperNVD in Section~\ref{sec:data}.

\subsection{Architectures}
\label{sec:archi}

The details of the architecture are presented in Table~\ref{tab:network_sup}. As explained in the main paper, our network has three main components: 
\begin{enumerate}[(1)]

\item The MAE encoder reduces the output of VideoMAE from shape $(\#videos, 768, 1568)$ to a compressed embedding shape of $(\#videos, 768, 1)$ using 1-linear-layer encoder. To guarantee the compressed embedding maintains VideoMAE's main features, the encoder is learned in combination with a decoder composed of a 1-linear layer that returns to the initial embedding shape $(\#videos, 768, 1568)$.

\item The main NVD model, whose architecture is shown in Figure 2 from the main paper, includes an alpha module and two Layer Modules (LM), each consisting of a mapping module, a texture module, and a residual module. Each of these modules is a 4-layer MLP (input layer, 2 hidden layers, output layer). The mapping module has an inner hidden dimension of 256 while all the rest have a hidden dimension of 64.
\item Our hypernet comprises 33 MLPs, each corresponding to a structural layer of the main net. Each of these MLPs has an input dimension of 768, a hidden layer of size 128, and outputs the predicted weights for its respective layer in the NVD model. 
\end{enumerate}

\begin{table*}[h]
\centering
\caption{Architecture details of our network. b stands for background, f stands for foreground.}
\begin{tabular}{cc|c|c|c|c|c}
\hline
Network &
Module &
  \begin{tabular}[c]{@{}c@{}}Number of nets\end{tabular} &
  \begin{tabular}[c]{@{}c@{}}Input dim\end{tabular} &
  \begin{tabular}[c]{@{}c@{}}Output dim\end{tabular} &
  \begin{tabular}[c]{@{}c@{}}Hidden layers\end{tabular} &
  \begin{tabular}[c]{@{}c@{}}Hidden dim\end{tabular} \\ \hline
\multirow{2}{*}{\begin{tabular}[c]{@{}c@{}}MAE\\ encoder\end{tabular}} & Encoder  & 1       & (\#videos, 768, 1568)   & (\#videos, 768, 1)      & 0 & --- \\
                                                                       & Decoder  & 1       & (\#videos, 768, 1)      & (\#videos, 768, 1568)   & 0 & --- \\ \hline
\multirow{4}{*}{\begin{tabular}[c]{@{}c@{}}NVD\\ model\end{tabular}}   & Mapping  & 2 (b\&f) & (\#points, 3) &  (\#points, 2) & 3 & 256 \\
                                                                       & Texture  & 2 (b\&f) & (\#points, 2) &  (\#points, 3) & 3 & 64  \\
                                                                       & Residual & 2 (b\&f) & (\#points, 3) &  (\#points, 1) & 3 & 64  \\
                                                                       & Alpha    & 1       &  (\#points, 3) &  (\#points, 1) & 3 & 64  \\ \hline
Hypernet                                                               & 1 Module & 33      & (\#videos, 768)& (\#videos, \#weights) & 2 & 128 \\ \hline
\end{tabular}%
\label{tab:network_sup}
\end{table*}

\subsection{Training losses}
\label{sec:loss}

The training objective of our Hypernet is the same as that of the previous neural video decomposition (NVD) models. Through these losses, we learn high-quality video reconstruction in a self-supervised manner. For this reason, we consider loss functions derived from previous research~\cite{kasten2021layered, chan2023hashing}. We adhere to the same notation as outlined in the main paper. These losses include:

\vspace{3mm}
    
\noindent \textbf{Reconstruction loss.} The reconstruction loss serves as the primary objective for self-supervised learning, ensuring the quality of video reconstruction. This loss comprises two terms: one addressing the RGB values and the other focusing on gradients,

\begin{equation}
\mathcal{L}_{recon}=\lambda_{r}\mathcal{L}_{rgb}+\lambda_{g}\mathcal{L}_{grad}.
\end{equation}
Here, $\lambda_{r}$ and $\lambda_{g}$ represent the weights, and are set to 5 and 1, respectively. The RGB term calculates the squared distance between the reconstructed color, $\hat{c}^p$, and the ground truth color, $c^p$, resulting in:

\begin{equation}
    \mathcal{L}_{rgb}=\left \|\hat{c}^p - c^p \right \| ^2_2 ,
\end{equation}
and the gradient term $\mathcal{L}_{grad}$ is given by:
\begin{equation}
\mathcal{L}_{grad} = \left \|\hat{d}_x- d_{x}\right \| ^2_2 +  \left \|\hat{d}_y-d_{y}\right \| ^2_2 ,
\end{equation}
where $(\hat{d}_x, \hat{d}_y)$ and $(d_x, d_y)$ are the spatial derivatives of the reconstructed image and the ground truth image, respectively.

\vspace{3mm}

\noindent \textbf{Consistency loss.} This loss ensures accurate motion representation supervised by optical flow. We expect corresponding pixels across the video to be mapped to the same point in the texture layer. For a pixel coordinate $p = (x, y, t)$, the corresponding point $p' = (x', y', t \pm 1)$, propagated using pre-computed forward or backward optical flow, is calculated as:

\begin{equation}
\begin{aligned}
\mathcal{L}_{flow-p}=&\alpha^{p}\left \|\mathcal{M}_{f}(p)-\mathcal{M}_{f}(p')\right \| \\ 
&+(1-\alpha^{p})\left \|\mathcal{M}_{b}(p)-\mathcal{M}_{b}(p')\right \|,
\end{aligned}
\end{equation}
where $\mathcal{M}_f$ and $\mathcal{M}_b$ are the foreground and background mapping functions, which map $p$ to the texture map coordinates $(u^p_f,v^p_f)$ and $(u^p_b,v^p_b)$, respectively. $\alpha^p$ represents the predicted opacity value of $p$. 

We also want the corresponding pixels to have the same alpha value:

\begin{equation}
\mathcal{L}_{flow-\alpha}=|\alpha^p-\alpha^{p'}|.
\end{equation}
Then, the total consistency loss is given by:

\begin{equation}
\mathcal{L}_{flow}=w^p(\lambda_{fp}\mathcal{L}_{flow-p}+\lambda_{f\alpha}\mathcal{L}_{flow-\alpha}),
\end{equation}
where we set $\lambda_{fp}=0.01$ and $\lambda_{f\alpha}=0.05$. $w^p$ indicates whether the correspondence between the two points $p$ and $p'$ is consistent, based on the standard forward-backward flow consistency check. Specifically, $w^p=1$ denotes consistent correspondence. To verify the reliability of the predicted optical flows, we perform a cycle mapping using the forward and backward flows. If the coordinate difference is smaller than 1 pixel, we consider it a reliable prediction and assign $w^p=1$; otherwise, we set $w^p=0$.

\vspace{3mm}

\noindent \textbf{Sparsity loss.} The sparsity loss is designed to prevent duplicate content across different texture layers.

\begin{equation}
\mathcal{L}_{sparsity}=\lambda_{s}\left \|(1-\alpha^p) c^p_f\right \|^2 ,
\end{equation}
where $c^p_f$ represents the predicted color at position $p$ for the foreground layer. We set $\lambda_{s}=1$ in the experiments.

\vspace{3mm}

\noindent \textbf{Residual consistency loss.} This loss ensures smooth lighting conditions while preventing the residual estimator from capturing all color details. Formally, it is defined as:

\begin{equation}
\mathcal{L}_{res}= \lambda_{res-s}\mathcal{L}_{smooth}+\lambda_{res-r}R_{res}, 
\end{equation}
where $\lambda_{res-s}=0.1$, $\lambda_{res-r}=0.5$.

First, the loss ensures smooth lighting conditions by constraining the values at the same position on the texture coordinates to be positively correlated. For a small $k \times k$ patch at time $t_1$ and $t_2$ in the residuals $r^p$, we define the residual consistency loss using normalized cross-correlation as:
\begin{equation}
\mathcal{L}_{smooth}= \cfrac{(r^p_{t_1}-\mu_{r^p_{t_1}})(r^p_{t_2}-\mu _{r^p_{t_2}})}{\sigma _{r^p_{t_1}}\sigma _{r^p_{t_2}}} + {\sigma}^2_{r^p_{t_2}},
\end{equation}
where $\mu$ and $\sigma$ are the mean and standard deviation of the corresponding patch, respectively. ${\sigma}^2_{r^p_{t_2}}$ is a variance-smoothness term. We set $k$ as 3.

Since only lighting changes are expected to be included in the residuals, a regularization term is applied to prevent the residual estimator $\mathcal{R}$ from capturing color details:
\begin{equation}
R_{res}= \left \| \mathcal{R}(\cdot )-1 \right \|.
\end{equation}

\noindent \textbf{Alpha regularization loss.} This additional regularization term ensures that the opacity map for each layer is clean and reliable, with lighting conditions properly embedded in each layer. It is defined using a BCE loss applied to the maximum values across all opacity maps:
\begin{equation}
\mathcal{L}_{\alpha reg}= \lambda_{\alpha reg}BCE(\underset{n \in \{0,...,N\}}{max} \alpha_n),
\end{equation}
where we choose $\lambda_{\alpha reg}=0.1$.

\vspace{4mm}

\noindent Two extra losses are applied just in the initial iterations:

\vspace{3mm}

\noindent \textbf{Rigidity loss.} The rigidity loss is designed to prevent the texture layer from degrading into a simple color palette or developing a distorted layout. It enforces local rigidity in the mapping from pixel locations in the video to the texture layer. This loss is applied to both foreground and background mappings:

\begin{equation}
\mathcal{L}_{rigid}=\lambda_{r}(D(\mathcal{M}_b)+D(\mathcal{M}_f)),
\end{equation}
where $\lambda_{r}=0.001$, and the loss is applied only in the first 5,000 iterations. For a giving mapping $\mathcal{M}$, the rigidity term $D$ is defined as a variant of symmetric Dirichlet term:
\begin{equation}
D(\mathcal{M})=\left \| J^T_\mathcal{M}J_\mathcal{M}\right \|_F + \left \| (J^T_\mathcal{M}J_\mathcal{M})^{-1}\right \|_F ,
\end{equation}
where $J_{\mathcal{M}}$ is the Jacobian matrix of $\mathcal{M}$, it is given by:
\begin{equation}
J_{\mathcal{M}}=\left [ \mathcal{M}(p_x) - \mathcal{M}(p) ~~~ \mathcal{M}(p_y) - \mathcal{M}(p) \right ],
\end{equation}
where $p_x=(x+1, y, t), p_y=(x, y+1, t)$.

\vspace{3mm}
    
\noindent \textbf{Alpha bootstrapping loss.} This loss provides initial guidance for the alpha module to predict a reasonable opacity map for layer separation. We compute the binary cross entropy loss between the expected alpha value $\alpha^p$ of the pixel $p$ and the corresponding value $m^p$ in the reference mask.

\begin{equation}
\mathcal{L}_{\alpha boot}=\lambda_{\alpha}BCE(\alpha^p, m^p)
\end{equation}

We set $\lambda_{\alpha}=2$ and apply the alpha bootstrapping loss during the first 10,000 iterations.
    
\vspace{4mm}

\noindent \textbf{Final loss.} The addition of the above-mentioned losses gives the total loss:
\begin{equation}
\mathcal{L}_{total} = \mathcal{L}_{recon}+\mathcal{L}_{flow}+\mathcal{L}_{sparsity}+\mathcal{L}_{res}+\mathcal{L}_{\alpha reg}
\end{equation}

\subsection{Training dataset}
\label{sec:data}

In Table~\ref{tab:video_sup}, we list the videos from the DAVIS dataset~\cite{perazzi2016benchmark} used for training the models presented in Table 2 of the main paper. Please recap that Table 2 of the main paper demonstrate the capability of our HyperNVD to train with different numbers of videos without compromising largely the quality reconstruction.  

\begin{table*}[h]
\centering
\caption{Videos used for training our metamodel in Table 2 of the main paper.}
\label{tab:video_sup}
\begin{tabular}{c|l}
\hline
Number &
  \multicolumn{1}{c}{Videos} \\ \hline
1 &
  \colorbox{blue!30}{hike} \\
3 &
  \colorbox{blue!30}{hike}, \colorbox{green!30}{car-turn, blackswan} \\ 
5 &
  \colorbox{blue!30}{hike}, \colorbox{green!30}{car-turn, blackswan}, \colorbox{purple!30}{bear, lucia} \\ 
10 &
  \begin{tabular}[c]{@{}l@{}}\colorbox{blue!30}{hike}, \colorbox{green!30}{car-turn, blackswan}, \colorbox{purple!30}{bear, lucia}, \colorbox{red!30}{boat, bus, kite-walk, rollerblade, stroller} \end{tabular} \\ 
15 &
  \begin{tabular}[c]{@{}l@{}}\colorbox{blue!30}{hike}, \colorbox{green!30}{car-turn, blackswan}, \colorbox{purple!30}{bear, lucia}, \colorbox{red!30}{boat, bus, kite-walk, rollerblade, stroller}, \\ \colorbox{yellow!30}{camel, car-roundabout, elephant, motorbike, train} \end{tabular} \\ 
30 &
  \begin{tabular}[c]{@{}l@{}}\colorbox{blue!30}{hike}, \colorbox{green!30}{car-turn, blackswan}, \colorbox{purple!30}{bear, lucia}, \colorbox{red!30}{boat, bus, kite-walk, rollerblade, stroller}, \\ \colorbox{yellow!30}{camel, car-roundabout, elephant, motorbike, train}, \\ \colorbox{orange!30}{bmx-bumps, bmx-trees, cow, drift-straight, flamingo, goat, horsejump-low, kite-surf, libby}, \\ \colorbox{orange!30}{paragliding,  scooter-gray, soapbox, soccerball, surf, swing} \end{tabular} \\ \hline

\end{tabular}%
\end{table*}

\section{More qualitative results}
\label{sec: quality}

In this section, we present more visualization results of our approach to video decomposition and editing. A video demo is also provided in the supplementary materials to intuitively showcase the decomposition and editing results.

\begin{figure*}[ht]
    \centering
    \includegraphics[width=1\textwidth]{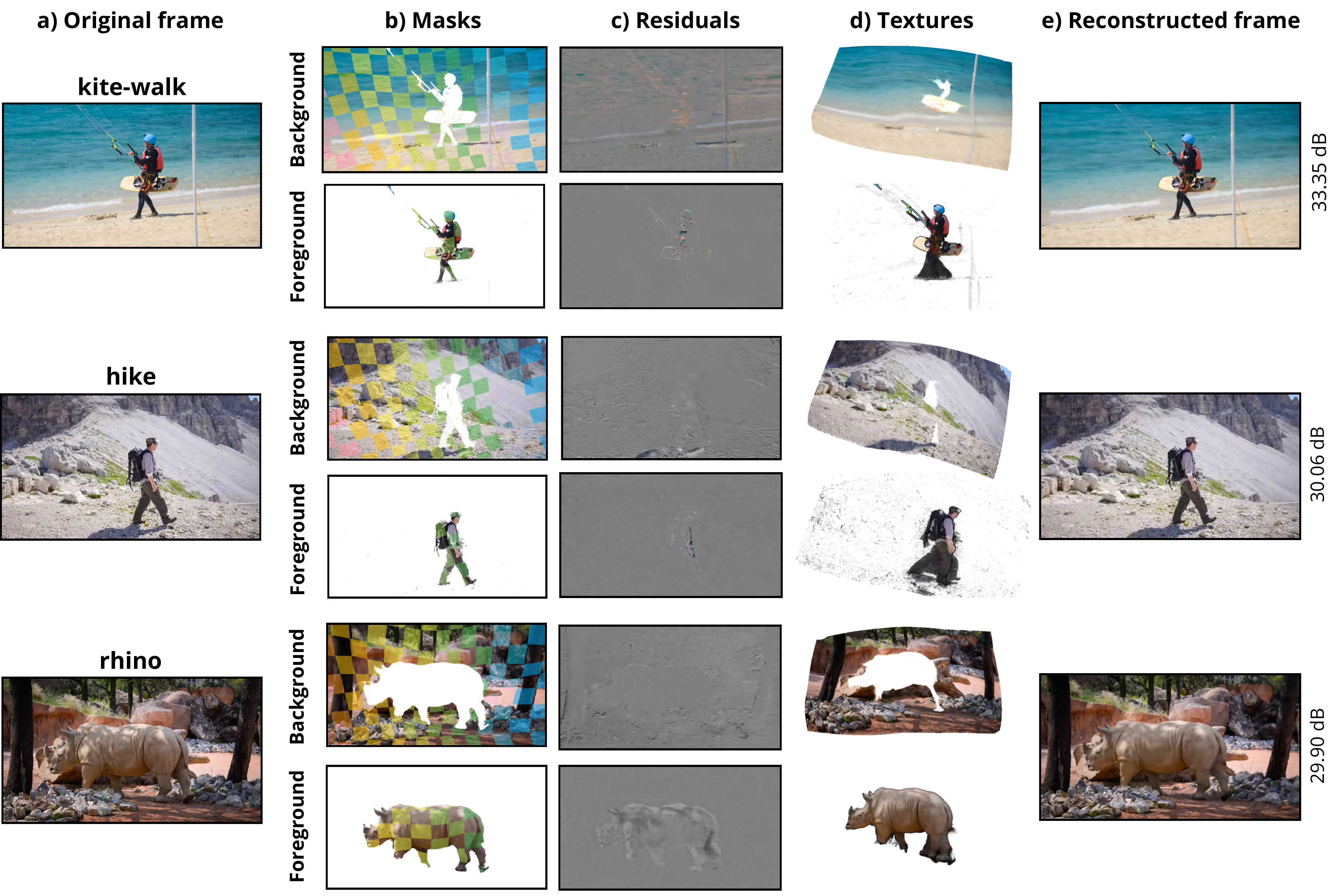}
    \caption{Qualitative results on the DAVIS dataset. Displayed are the a) original frames, b) predicted masks, c) residual maps, d) texture maps, e) reconstructed frames, and PSNR values for different videos. A color checkerboard overlay is applied to the masked areas to visualize texture transformations.}
    \label{fig:decompose_sup}
\end{figure*}

Figure~\ref{fig:decompose_sup} illustrates the video decomposition results. The three examples in the figure achieve a PSNR of approximately 30 dB and effectively differentiate between the foreground and background. Notably, the shadows on the rhino's back are accurately decomposed into the residual map. 

\begin{figure*}
    \centering
    \includegraphics[width=1\textwidth]{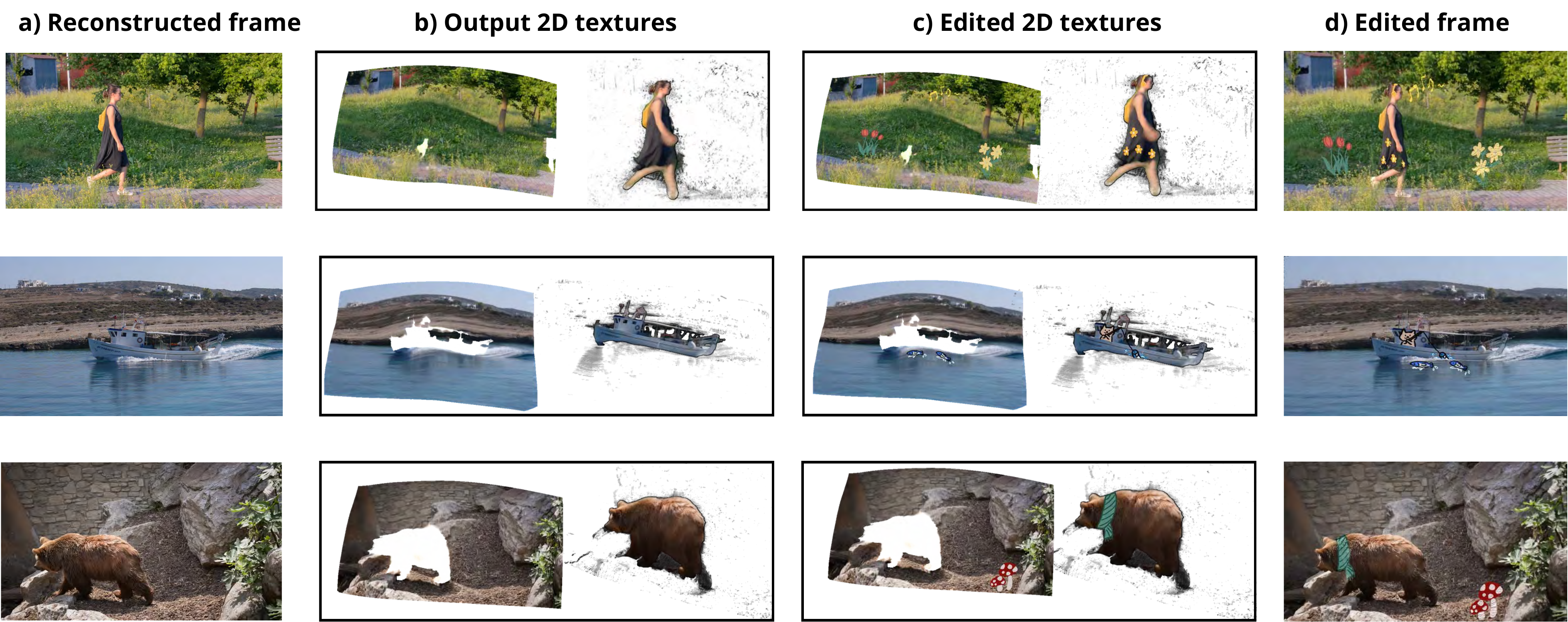}
    \caption{Edited video results. Displayed are the a) reconstructed frame with our HyperNVD, b) 2D texture maps, c) edited texture maps, and d) edited frame. We add doodles to both the foreground and background texture layers. The edited video maintains good visual quality and smooth transitions.}
    \label{fig:edit_sup}
\end{figure*}

Figure~\ref{fig:edit_sup} demonstrates the video editing results. By modifying the foreground and background texture maps, we observe that elements added to the foreground ---such as the bear's scarf and the headphones and flowers on Lucia's skirt--- naturally adapt to the motion of the foreground. For these videos, where the backgrounds remain relatively stable, elements added to the background exhibit no significant motion and blend naturally.

Please note that in all current models ---including ours--- editing is restricted to non-transparent areas of the texture map due to the requirement of alpha-weighted compositing for blending layers. Edits applied to transparent regions will not appear in the final edited video. To add elements outside the object's region that move along with the object, it is necessary to modify the opacity map. This limitation is inherent to video layer-based editing and is also present in previous methods. Despite this constraint, we believe that addressing such challenges could open avenues for future research and enable exciting new applications.

\section{Limitation and discussion}

In terms of limitations of our work, since our hypernetwork uses an MLP to predict parameters for each layer of the target network, resulting in a large parameter count, further structural optimization and techniques like Low-Rank Factorization \cite{skorokhodov2021adversarial} could help reduce the hypernetwork's parameter size.

We have made some interesting observations. For example, using different initial masks does not significantly affect the final reconstruction quality, but finer masks guide the network to focus on object details. Therefore, in practical applications where precise editing of edges or related effects is required, using more defined masks is advisable. These can be improved with the latest segmentation methods, such as SAM2 \cite{ravi2024sam}.

\end{document}